\documentclass[runningheads]{llncs}
\usepackage{graphicx}
\usepackage{hyperref}

\usepackage[usenames,dvipsnames]{xcolor}
\usepackage{listings}

\definecolor{light-gray}{gray}{0.99}
\usepackage{tikz}

\begin{document}
%
\title{A Showcase of the Use of Autoencoders in Feature Learning Applications\thanks{D. Charte is supported by by the Spanish Ministry of Science, Innovation and Universities under the FPU National Program (Ref. FPU17/04069). This work is supported by the Spanish National Research Projects TIN2015-68854-R and TIN2017-89517-P.}}
%
%
\author{David Charte\inst{1}\orcidID{0000-0002-4830-9512} \and
Francisco Charte\inst{2}\orcidID{0000-0002-3083-8942} \and
Mar{\'i}a J. del Jesus\inst{2}\orcidID{0000-0002-7891-3059} \and
Francisco Herrera\inst{1}\orcidID{0000-0002-7283-312X}}
\authorrunning{D. Charte et al.}
%
\institute{Andalusian Research Institute in Data Science and Computational Intelligence, Dept. of Computer Science and Artificial Intelligence, Universidad de Granada \\ Periodista Daniel Saucedo Aranda, s/n, 18071 Granada, Spain \\
\email{fdavidcl@ugr.es,herrera@decsai.ugr.es} \and
Andalusian Research Institute in Data Science and Computational Intelligence, Computer Science Dept., Universidad de Ja\'en \\ Campus Las Lagunillas, s/n, 23071 Ja\'en, Spain 
\\
\email{\{fcharte,mjjesus\}@ujaen.es}}
\maketitle              
\begin{abstract}

Autoencoders are techniques for data representation learning based on artificial neural networks. Differently to other feature learning methods which may be focused on finding specific transformations of the feature space, they can be adapted to fulfill many purposes, such as data visualization, denoising, anomaly detection and semantic hashing. 

This work presents these applications and provides details on how autoencoders can perform them, including code samples making use of an R package with an easy-to-use interface for autoencoder design and training, \texttt{ruta}. Along the way, the explanations on how each learning task has been achieved are provided with the aim to help the reader design their own autoencoders for these or other objectives. 

This manuscript was accepted as conference paper in IWINAC 2019. The final authenticated publication is available online at \url{https://doi.org/10.1007/978-3-030-19651-6_40}.

\end{abstract}
\section{Introduction}

Autoencoders (AEs) \cite{hinton} are versatile unsupervised learning methods. Also known as autoassociators, since their first uses their purpose has usually been to transform the input variable space into a more useful one, either one which is more compact, or whose structure is simpler or more convenient. In the last decade, deep learning methods have been gaining use since computing capabilities and optimization methods have improved. As a consequence, the topic of representation learning \cite{bengio} has attained more interest, and AEs with it.

Nowadays, many variants of AEs \cite{charte} have been developed with several applications in mind. These cover from simple dimensionality reduction to instance generation, data enhancement and hashing. Furthermore, they are not limited to problems with a standard (input, label) structure, but AE models have been defined for many nonstandard learning problems such as multi-view learning or multi-label learning \cite{nonstandard,advcae,maniac}.

In the following sections, AEs as a feature learning tool are described and several different applications are detailed and illustrated with examples. Each example comprises several lines of code which tackle a given task and some graphical output. The complete code samples are available at \url{https://github.com/ari-dasci/autoencoder-showcase}.

This document is structured as follows. Section~\ref{sec.fundamentals} introduces the inner workings of AEs. Section~\ref{sec.visualization} shows a simple visualization task, followed by an example on image denoising in Section~\ref{sec.denoising}. Afterwards, Section~\ref{sec.anomaly} describes how to detect anomalous samples with AEs and Section~\ref{sec.hashing} a way to perform semantic hashing. Last, Section~\ref{sec.other} mentions other possible applications of AEs, and Section~\ref{sec.conclusions} draws some conclusions.

\section{Fundamentals of autoencoders}
\label{sec.fundamentals}

AEs are artificial neural networks designed to find an alternative representation of data with some desirable properties. They are generally unsupervised techniques, that is, AEs are usually intended to complete their task without any class information. They are comprised of two distinct components: the encoder and the decoder. Both have interesting outputs: the first provides the encodings for each input instance, and the second obtains a reconstruction of the original instance. The basic objective for an AE consists in finding the parameters that allow to retrieve the most faithful reconstructions.

\begin{figure}[ht]
  \centering
  \resizebox {0.45\columnwidth} {!} {
        \begin{tikzpicture}[scale=0.12]
\tikzstyle{every node}+=[inner sep=0pt]
\draw [black] (15.7,-15.6) circle (3);
\draw [black] (15.7,-23.8) circle (3);
\draw [black] (15.7,-31.9) circle (3);
\draw [black] (15.7,-39.6) circle (3);
\draw [black] (15.7,-47.6) circle (3);
\draw [black] (29.3,-23.8) circle (3);
\draw [black] (29.3,-31.9) circle (3);
\draw [black] (29.3,-39.6) circle (3);
\draw [black] (41.8,-27) circle (3);
\draw [black] (41.8,-36.5) circle (3);
\draw [black] (53.7,-23.8) circle (3);
\draw [black] (53.7,-31.9) circle (3);
\draw [black] (53.7,-39.6) circle (3);
\draw [black] (65.4,-15.6) circle (3);
\draw [black] (65.4,-23.8) circle (3);
\draw [black] (65.4,-31.9) circle (3);
\draw [black] (65.4,-39.6) circle (3);
\draw [black] (65.4,-47.6) circle (3);
\draw [black] (18.27,-17.15) -- (26.73,-22.25);
\fill [black] (26.73,-22.25) -- (26.3,-21.41) -- (25.79,-22.27);
\draw [black] (17.62,-17.9) -- (27.38,-29.6);
\fill [black] (27.38,-29.6) -- (27.25,-28.66) -- (26.48,-29.3);
\draw [black] (17.18,-18.21) -- (27.82,-36.99);
\fill [black] (27.82,-36.99) -- (27.86,-36.05) -- (26.99,-36.54);
\draw [black] (18.7,-23.8) -- (26.3,-23.8);
\fill [black] (26.3,-23.8) -- (25.5,-23.3) -- (25.5,-24.3);
\draw [black] (18.28,-25.34) -- (26.72,-30.36);
\fill [black] (26.72,-30.36) -- (26.29,-29.53) -- (25.78,-30.39);
\draw [black] (17.66,-26.07) -- (27.34,-37.33);
\fill [black] (27.34,-37.33) -- (27.2,-36.39) -- (26.44,-37.05);
\draw [black] (18.28,-30.36) -- (26.72,-25.34);
\fill [black] (26.72,-25.34) -- (25.78,-25.31) -- (26.29,-26.17);
\draw [black] (18.7,-31.9) -- (26.3,-31.9);
\fill [black] (26.3,-31.9) -- (25.5,-31.4) -- (25.5,-32.4);
\draw [black] (18.31,-33.38) -- (26.69,-38.12);
\fill [black] (26.69,-38.12) -- (26.24,-37.29) -- (25.75,-38.16);
\draw [black] (17.66,-37.33) -- (27.34,-26.07);
\fill [black] (27.34,-26.07) -- (26.44,-26.35) -- (27.2,-27.01);
\draw [black] (18.31,-38.12) -- (26.69,-33.38);
\fill [black] (26.69,-33.38) -- (25.75,-33.34) -- (26.24,-34.21);
\draw [black] (18.7,-39.6) -- (26.3,-39.6);
\fill [black] (26.3,-39.6) -- (25.5,-39.1) -- (25.5,-40.1);
\draw [black] (17.19,-45) -- (27.81,-26.4);
\fill [black] (27.81,-26.4) -- (26.98,-26.85) -- (27.85,-27.35);
\draw [black] (17.66,-45.33) -- (27.34,-34.17);
\fill [black] (27.34,-34.17) -- (26.43,-34.44) -- (27.19,-35.1);
\draw [black] (18.29,-46.08) -- (26.71,-41.12);
\fill [black] (26.71,-41.12) -- (25.77,-41.1) -- (26.28,-41.96);
\draw [black] (32.21,-24.54) -- (38.89,-26.26);
\fill [black] (38.89,-26.26) -- (38.24,-25.57) -- (37.99,-26.54);
\draw [black] (32.09,-30.81) -- (39.01,-28.09);
\fill [black] (39.01,-28.09) -- (38.08,-27.92) -- (38.44,-28.85);
\draw [black] (31.41,-37.47) -- (39.69,-29.13);
\fill [black] (39.69,-29.13) -- (38.77,-29.35) -- (39.48,-30.05);
\draw [black] (31.4,-25.94) -- (39.7,-34.36);
\fill [black] (39.7,-34.36) -- (39.49,-33.44) -- (38.78,-34.14);
\draw [black] (32.12,-32.94) -- (38.98,-35.46);
\fill [black] (38.98,-35.46) -- (38.41,-34.72) -- (38.06,-35.66);
\draw [black] (32.21,-38.88) -- (38.89,-37.22);
\fill [black] (38.89,-37.22) -- (37.99,-36.93) -- (38.23,-37.9);
\draw [black] (44.7,-37.26) -- (50.8,-38.84);
\fill [black] (50.8,-38.84) -- (50.15,-38.16) -- (49.9,-39.13);
\draw [black] (44.6,-35.42) -- (50.9,-32.98);
\fill [black] (50.9,-32.98) -- (49.98,-32.8) -- (50.34,-33.74);
\draw [black] (43.85,-34.31) -- (51.65,-25.99);
\fill [black] (51.65,-25.99) -- (50.74,-26.23) -- (51.47,-26.91);
\draw [black] (44.7,-26.22) -- (50.8,-24.58);
\fill [black] (50.8,-24.58) -- (49.9,-24.3) -- (50.16,-25.27);
\draw [black] (44.57,-28.14) -- (50.93,-30.76);
\fill [black] (50.93,-30.76) -- (50.38,-29.99) -- (50,-30.92);
\draw [black] (43.86,-29.18) -- (51.64,-37.42);
\fill [black] (51.64,-37.42) -- (51.45,-36.49) -- (50.73,-37.18);
\draw [black] (56.16,-22.08) -- (62.94,-17.32);
\fill [black] (62.94,-17.32) -- (62,-17.37) -- (62.58,-18.19);
\draw [black] (56.7,-23.8) -- (62.4,-23.8);
\fill [black] (62.4,-23.8) -- (61.6,-23.3) -- (61.6,-24.3);
\draw [black] (56.17,-25.51) -- (62.93,-30.19);
\fill [black] (62.93,-30.19) -- (62.56,-29.33) -- (61.99,-30.15);
\draw [black] (56.21,-33.55) -- (62.89,-37.95);
\fill [black] (62.89,-37.95) -- (62.5,-37.09) -- (61.95,-37.93);
\draw [black] (55.49,-26.21) -- (63.61,-37.19);
\fill [black] (63.61,-37.19) -- (63.54,-36.25) -- (62.74,-36.84);
\draw [black] (55.02,-26.49) -- (64.08,-44.91);
\fill [black] (64.08,-44.91) -- (64.17,-43.97) -- (63.27,-44.41);
\draw [black] (55.45,-29.46) -- (63.65,-18.04);
\fill [black] (63.65,-18.04) -- (62.78,-18.4) -- (63.59,-18.98);
\draw [black] (56.17,-30.19) -- (62.93,-25.51);
\fill [black] (62.93,-25.51) -- (61.99,-25.55) -- (62.56,-26.37);
\draw [black] (56.7,-31.9) -- (62.4,-31.9);
\fill [black] (62.4,-31.9) -- (61.6,-31.4) -- (61.6,-32.4);
\draw [black] (55.49,-34.31) -- (63.61,-45.19);
\fill [black] (63.61,-45.19) -- (63.53,-44.25) -- (62.73,-44.85);
\draw [black] (55.01,-36.9) -- (64.09,-18.3);
\fill [black] (64.09,-18.3) -- (63.29,-18.8) -- (64.18,-19.23);
\draw [black] (55.49,-37.19) -- (63.61,-26.21);
\fill [black] (63.61,-26.21) -- (62.74,-26.56) -- (63.54,-27.15);
\draw [black] (56.21,-37.95) -- (62.89,-33.55);
\fill [black] (62.89,-33.55) -- (61.95,-33.57) -- (62.5,-34.41);
\draw [black] (56.7,-39.6) -- (62.4,-39.6);
\fill [black] (62.4,-39.6) -- (61.6,-39.1) -- (61.6,-40.1);
\draw [black] (56.18,-41.29) -- (62.92,-45.91);
\fill [black] (62.92,-45.91) -- (62.55,-45.04) -- (61.98,-45.87);
\end{tikzpicture}

  }
  \caption[Sample autoencoder]{Typical organization of neurons inside an autoencoder, in this case, for a 2-variable encoding.}
  \label{fig:ae-example}
\end{figure}
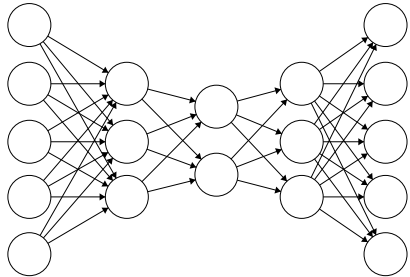

Fig.~\ref{fig:ae-example} shows the architecture of a simple AE, where input data is feeded into the leftmost layer and the obtained output must match it with as little error as possible. This can be interpreted as a parametrized mapping $f_{\theta}$ which is fitted to the data by minimizing a loss function $\mathcal J(\theta) = \sum_x L(x,f_{\theta}(x))$.

A common application of these models is transforming the variable space into a lower-dimensional one, while keeping enough information about samples to accurately recover their values in the original variable space. In this case, AEs can be seen as a nonlinear generalization of principal components analysis \cite{ANNsPCA}. However, they can show much more potential when combined with certain restrictions or modifications.

\subsection{Encoded space structure}

AEs can be altered in order for the encoded variable space to have some desirable structure, or to verify some properties. Following are some viable structures that may be applied to an AE. Among other possibilities, the encoding space can:  

\begin{itemize}
    \item be sparse, i.e. few neurons have a high value for each instance (sparse AE)
    \item resist distortions present in input data (denoising AE, robust AE)
    \item preserve local behavior and relative distances from the input space (contractive AE)
    \item follow a given distribution (adversarial AE)
\end{itemize}


\subsection{Available software}

Any well known deep learning framework may serve as base for the construction of AEs, but very few tools facilitate the design process by abstracting common traits from usual AEs. One of them is R package \texttt{ruta} \cite{ruta}, which provides a variety of AE variants, allowing for easy definition, training and usage of these models. The package is available on CRAN and thus can be installed with:

\begin{lstlisting}[language=R]
install.packages("ruta")
\end{lstlisting}

The upcoming sections will make use of this software and other libraries in order to provide compact code listings which achieve the different tasks. 
Other available software packages for the purposes of autoencoder training are: \texttt{autoencoder} (R), \texttt{h2o} (multiplatform) and \texttt{yadlt} (Python).

\section{Examples of use}
\label{sec.examples}

The following sections contain four complete examples where an AE is used to complete a learning task. Each example is comprised of a description of the tackled problem, some explanation on how an AE solves or can help solve it, a code snippet which trains an AE adequate for the task and some visual results and associated analysis.

\subsection{Visualization of high-dimensional data} 
\label{sec.visualization}

AEs can be arranged so as to produce a two-dimensional or three-dimensional code. This way, encoded instances can be directly represented in a graph. If samples are labeled with classes or target values, this can be used to provide a visual intuition on how these labels are organized. 

In the following example, R packages \texttt{ruta}, \texttt{scatterplot3d} and \texttt{colorspace} are loaded. This allows to train an AE which compresses data to a three-dimensional space, extract codifications for the desired dataset and lastly plot them with colors according to their class. The chosen AE has several hidden layers in order to allow it to learn short encodings; the middle layer uses sigmoid activation which limits the codes to the $[0,1]$ interval.

\begin{lstlisting}[language=R]
network <- input() + dense(12, activation = "relu") + dense(3, 
  activation = "sigmoid") + dense(12, activation = "relu") + output("linear")
model <- autoencoder_sparse(network) %>% train(x_train, epochs = 40)
codes <- model %>% encode(x_train)
scatterplot3d(codes, color = rainbow_hcl(7)[class], (1:7)[class])
\end{lstlisting}

\begin{figure}[ht]
    \centering
    \includegraphics[width=.49\textwidth]{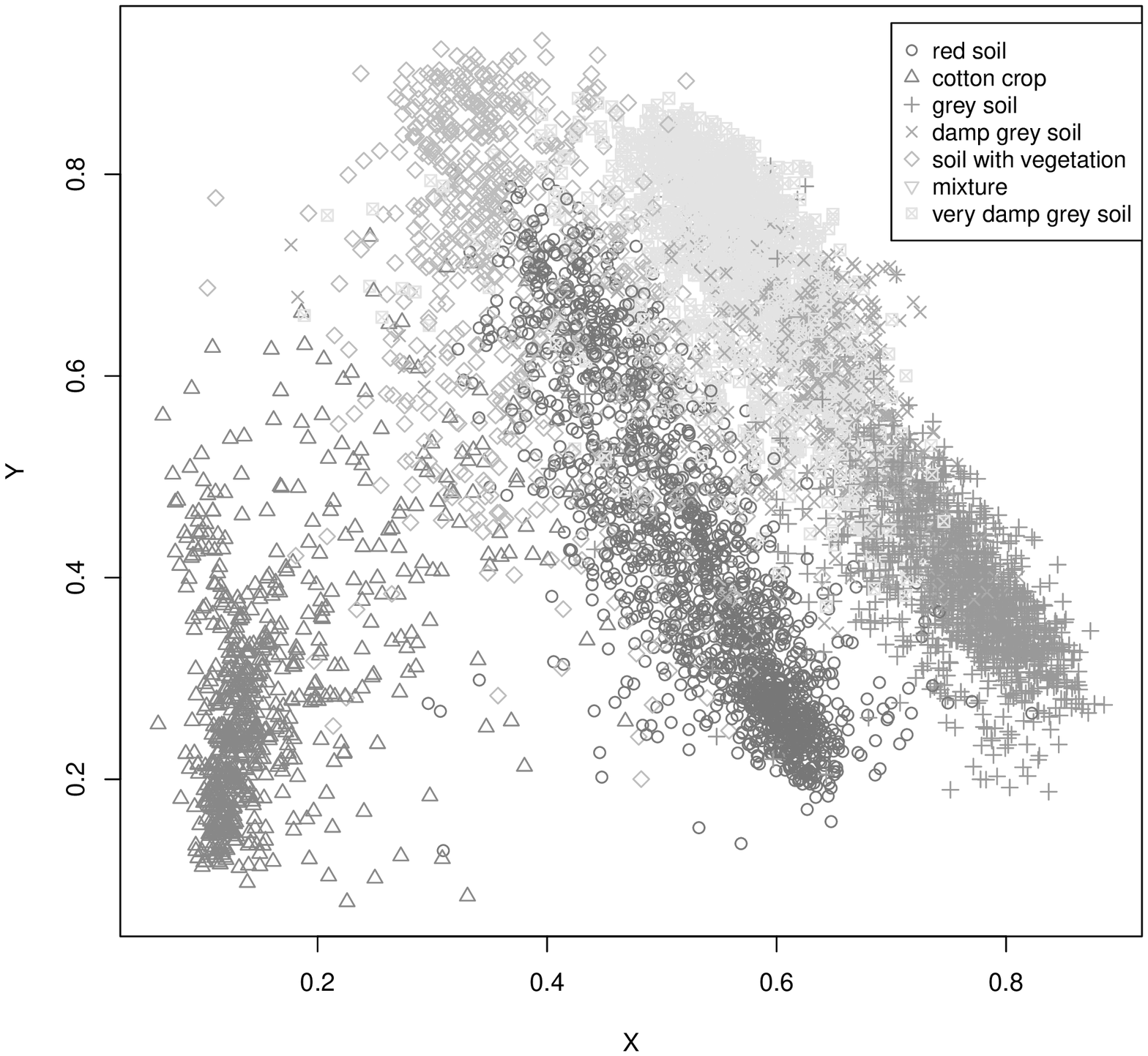}
    \includegraphics[width=.49\textwidth]{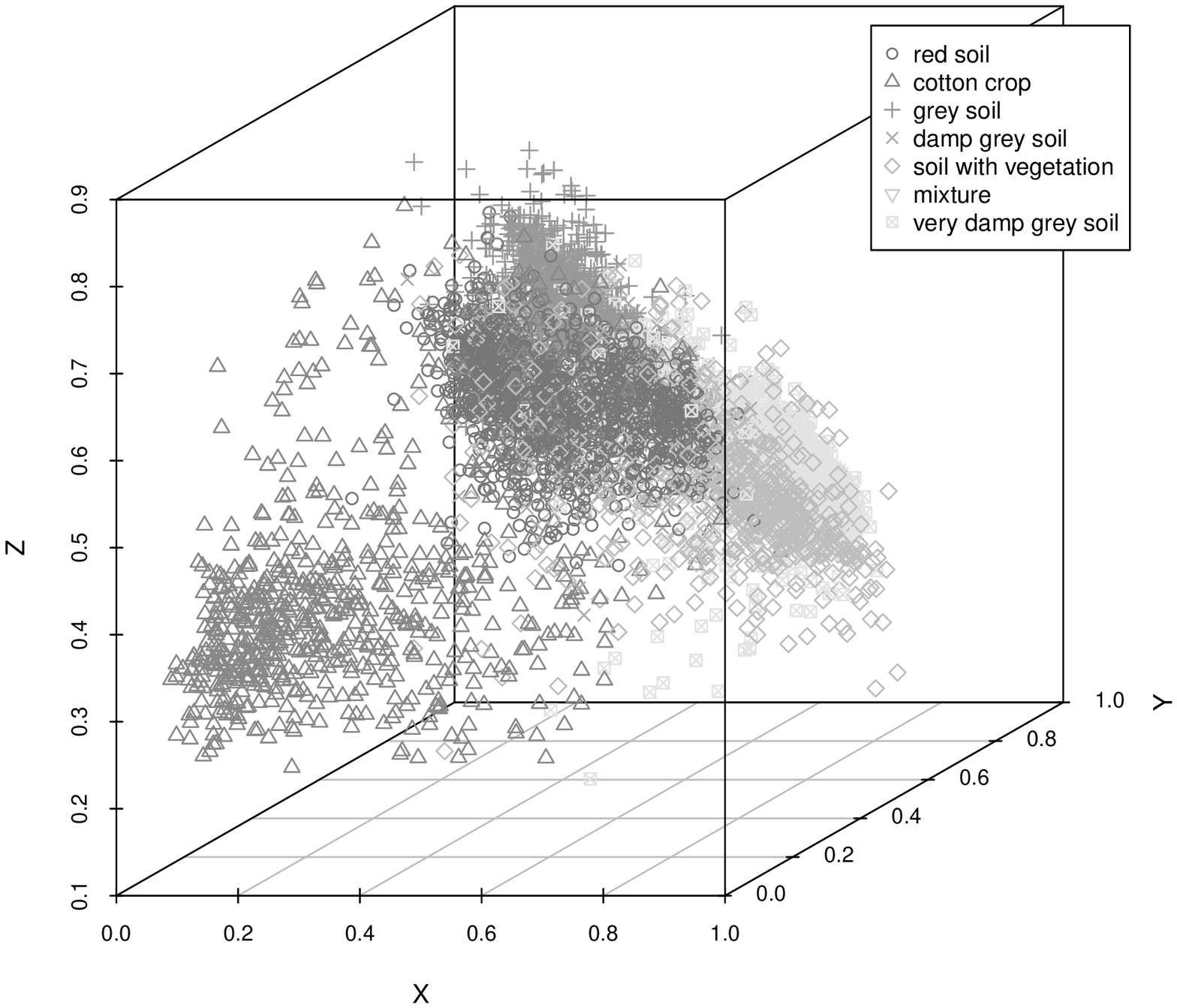}
    \caption{Two and three-dimensional projections of the 36-variable dataset Statlog.}
    \label{fig:visualization}
\end{figure}

Fig.~\ref{fig:visualization} shows the result of running this code on the 36-dimensional dataset Statlog (Landsat Satellite) \cite{ucirepo}, 
where each instance represents a small patch of land and is labeled with the type of land associated to the central pixel. Although the graph shows some fairly separated instance clusters, it is important to remark that the AE receives no class information and thus it can be assumed that the input features also present some degree of class separation.

\subsection{Image denoising}
\label{sec.denoising}

Since AEs are not only trained to transform the variable space but also to reconstruct the original one, a specific strategy can be used so that the reconstruction process eliminates distortions or noise \cite{xie}. AEs that are trained this way are usually called denoising AEs.

For the purposes of image denoising, a convolutional AE can be used. Bidimensional convolutional layers \cite{dlbook-conv} help models take advantage of the structure of images to reduce the number of parameters. The architecture of this AE does not need to reduce the dimension of the input data in the encoding layer. Instead, it is possible to increase the dimension in order for it to be able to discern useful information from noise. 

In this case, a very simple architecture has been used: one convolutional layer with an upsampling operation, another one with a max-pooling operation (to decrease the dimension) and an output convolutional layer with as many filters as channels in the data. The last layer uses a sigmoid activation because the inputs are normalized to the $[0,1]$ interval. This AE has been trained and tested with the CIFAR10 dataset included in Keras.

\begin{lstlisting}[language=R]
network <- input() + conv(16, 3, upsampling=2, activation="relu") + conv(16, 3, max_pooling=2, activation="relu") + conv(3, 3, activation="sigmoid")
model <- autoencoder_denoising(network, loss = "binary_crossentropy", 
  noise_type = "gaussian", sd = .05) %>%
  train(x_train, epochs = 30, batch_size = 500, optimizer = "adam")
noisy <- noise_gaussian(sd = .05) %>% apply_filter(x_test)
recovered <- model %>% reconstruct(noisy)
\end{lstlisting}

\begin{figure}[ht]
    \centering
    \includegraphics[width=\textwidth]{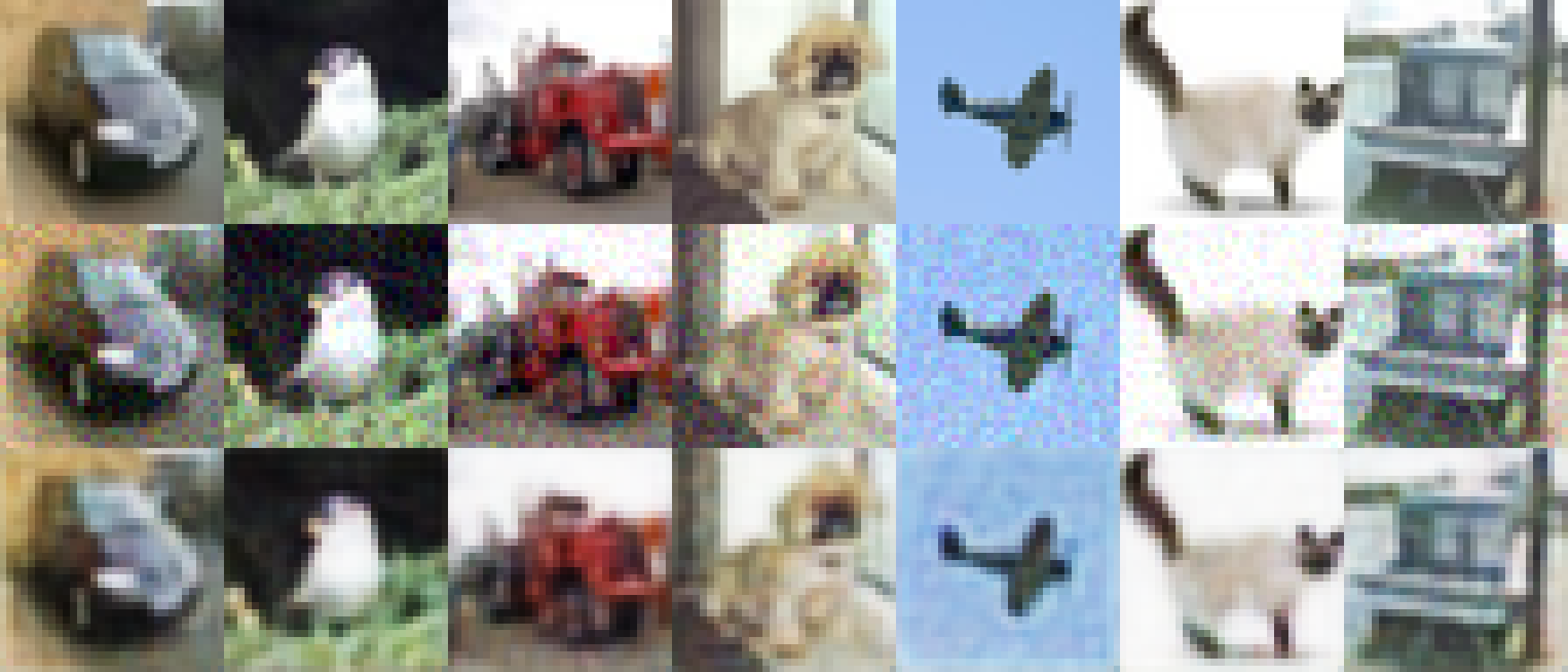}
    \caption{Image denoising. First row shows original test samples, second row displays the noisy images feeded to the AE, and third row shows reconstructed images.}
    \label{fig:denoising}
\end{figure}

The listing above also shows a way of introducing noise to the test subset and reconstructing it by means of the trained model. Fig.~\ref{fig:denoising} shows a sample of the obtained results drawn by means of the \texttt{grid} package: each test image, previously unseen by the AE, is joined by its noisy version, which is feeded to the trained model, and the reconstructed version, obtained at the output of the AE.

\subsection{Anomaly detection}
\label{sec.anomaly}

The encoded variable space generated by an AE usually omits information that is common to most or all instances, since the objective of an AE is to produce precise reconstructions while retaining only necessary information. Any traits present in all samples can be easily learned and recovered by the decoder, so that they do not need to be encoded. This means that, in the case that a new instance is very different from those of the training set, its reconstruction by the AE will predictably lose information and produce a high error. In these scenarios, AEs can serve as anomaly detectors \cite{sakurada,park}. This is especially useful when treating time series that may have abnormal regions, since detecting single examples based just on their reconstruction can be more challenging.

The following code trains a very simple denoising AE with a 16-variable encoding, and then measures the individual error made for each instance. The reason a denoising AE has been chosen in this case is that it can indirectly find lower dimensional manifolds in the data, and attempt to push instances to those when decoding. Thus, anomalous samples far from the manifold should stand out more when measuring the error.

\begin{lstlisting}[language=R]
model <- autoencoder_denoising(input() + dense(16, "sigmoid") + output()) %>% 
  train(x_train, epochs = 50, batch_size = 32)
errors <- rowMeans((model %>% reconstruct(x_test) - x_test) ** 2)
\end{lstlisting}

\begin{figure}[ht]
    \centering
    \includegraphics[width=.75\textwidth]{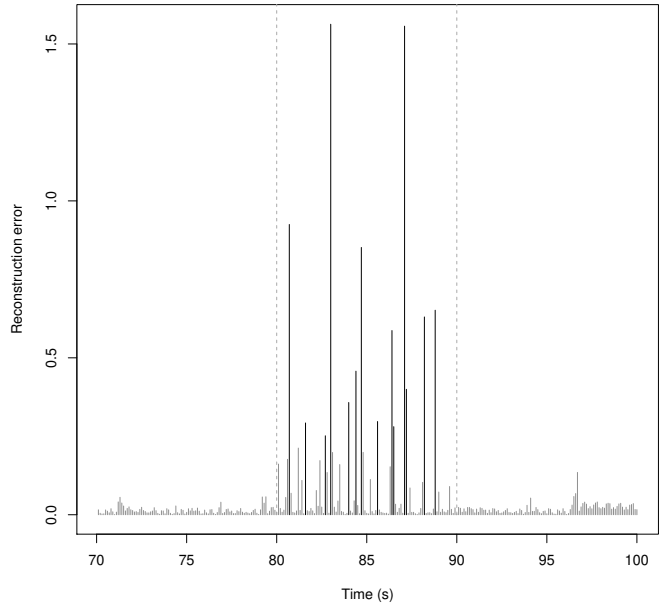}
    \caption{Reconstruction error of a test subset of a synthetic time series. Artificially generated anomalous data is placed between the dashed lines. High reconstruction errors (above the mean error plus its standard deviation) are marked in black.}
    \label{fig:anomaly}
\end{figure}

The AE above has been trained with a synthetic multi-valued time series based on samples from a solution to a Lorenz system, following the experimentation in \cite{sakurada}. An anomalous region has been artificially introduced within the test subset. The reconstruction error has been measured for each test instance, and the results are shown in the plot in Fig.~\ref{fig:anomaly}: anomalies present much higher errors in general, which can be used to detect that region.

\subsection{Semantic hashing}
\label{sec.hashing}

Semantic hashing \cite{hinton} is a task consisting on finding short binary codes for data points so that codes with small Hamming distance correspond to similar samples, and those separated by a high Hamming distance belong to very different points.

An AE can produce encodings in the range $[0,1]$ when the encoding layer has a sigmoid activation function. This can be joined with a high Gaussian noise at the input of that layer in order to strongly polarize its outputs \cite{hinton}. The results are very close to binary, and a thresholding function can be applied to obtain exact integers.

In the following example, the network defined includes a custom Keras layer which introduces this Gaussian noise and is otherwise very standard. A basic AE is trained using this network, afterwards it encodes test instances, and the \texttt{hash} function applies a threshold to each encoding to find a binary hash.

\begin{lstlisting}[language=R]
net <- input() + dense(256) + layer_keras("gaussian_noise", stddev = 16) +
  dense(10, activation = "sigmoid") + dense(256) + output("sigmoid")
model <- autoencoder(net, "binary_crossentropy") %>% 
  train(x_train, epochs = 50)
hash <- function(model, x, threshold = 0.5) {
  t(encode(model, x) %>% apply(1, function(r) as.integer(r > threshold)))
}
hashes <- model %>% hash(x_test)
\end{lstlisting}

In order to illustrate the effectiveness of this AE, we have trained it with the training subset of the IMDB dataset available in Keras, composed of 25\,000 movie reviews, from which just the 1\,000 most frequent words have been used (i.e. the training subset had 1\,000 variables and 25\,000 samples). Then, we use the obtained model to produce hashes for each test instance. 

\begin{figure}[ht]
    \centering
    \includegraphics[width=.75\textwidth]{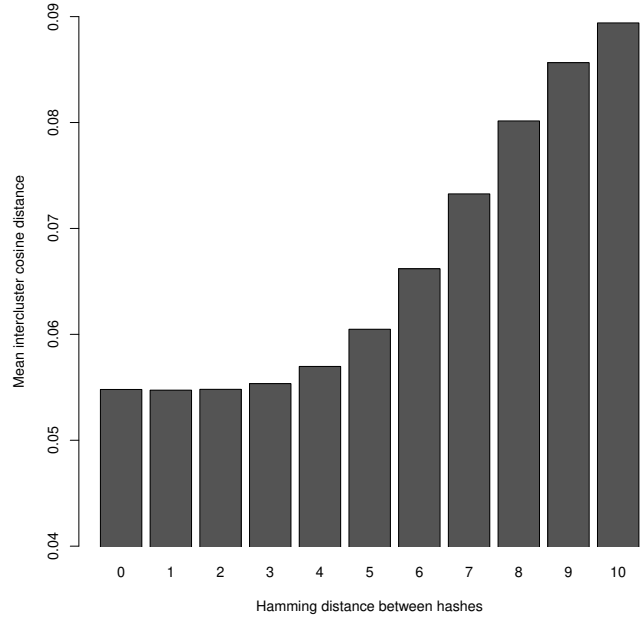}
    \caption{Measure of average cosine distance among instances whose hash encodings differ in any Hamming distance.}
    \label{fig:hashing}
\end{figure}

Since the objective of semantic hashing is to obtain similar binary hashes for similar inputs, we measure the distance among instances and associate it to that among their hashes. The dataset used contains text documents, so the cosine distance is a good measure for pairs of instances in this case. Hashes will be considered distant according to their Hamming distance. Fig.~\ref{fig:hashing} shows that, the more different two hashes are, more distance is between the corresponding instances. 

\subsection{Other applications}
\label{sec.other}

The applications described in the examples above are only a sample of the possible purposes an AE can be used for. Another common objective for AEs is to improve classification performance \cite{ssae,xu2017}. As with image denoising, AEs can serve as enhancers for other kinds of data; particularly speech \cite{speech} and electrocardiogram signals \cite{ecg}.

Combinations of AEs can merge information from several sources, such as two and three-dimensional human poses \cite{multimodal} or pairs of image and text and other multi-view data \cite{advcae}. Other AE-based models have been developed in order to improve tag recommendation \cite{rsdae}, prediction of movement from static images \cite{uncertain}, and translation involving sentence reordering \cite{translation}.

\section{Conclusions}
\label{sec.conclusions}

Feature learning is a crucial task which can determine the performance of a machine learning model. In this work, AEs have been described as an adaptable basis for many different tasks which involve finding alternative representations of data. 

Four example applications have been described in detail and solved using AEs: data visualization in two and three dimensions, denoising of images, detection of abnormal patterns in time series and semantic hashing for text documents. The sample experiments have been performed with well known datasets and using common features in AEs readily available in a published software package, \texttt{ruta}.

The main objective of this work has been to gather the necessary knowledge and ideas to guide the reader into designing their own experiments based on AE models when facing feature learning tasks.

\end{document}